# Value of Evidence on Influence Diagrams

Kazuo J. Ezawa
AT&T Bell Laboratories
600 Mountain Avenue
Murray Hill, NJ 07974

## Abstract

In this paper, we introduce evidence propagation operations on influence diagrams and a concept of value of evidence, which measures the value of experimentation. Evidence propagation operations are critical for the computation of the value of evidence, general update and inference operations in normative expert systems which are based on the influence diagram (generalized Bayesian network) paradigm. The value of evidence allows us to compute directly an outcome sensitivity, a value of perfect information and a value of control which are used in decision analysis (the science of decision making under uncertainty). More specifically, the outcome sensitivity is the maximum difference among the values of evidence, the value of perfect information is the expected value of the values of evidence, and the value of control is the optimal value of the values of evidence. We also discuss an implementation and a relative computational efficiency issues related to the value of evidence and the value of perfect information.

## 1. INTRODUCTION

Influence diagrams have been in use for decision analysis for the representation of decision problems as well as an decision evaluation tool. Evidence propagation on probabilistic influence diagrams which contain only chance (probabilistic) nodes has been previously discussed [Shachter 1990]. In this paper we discuss evidence propagation operations on influence diagrams and the value of evidence.

Probabilistic expert systems that can interact with and interpret real world quantitative data and uncertainty have been developed thanks to algorithmic advances on evidence propagation (probabilistic inference) on Bayesian networks [Shachter 1990&1992, Lauritzen 1988, Pearl 1988, Jensen 1990]. It replaces a rule-based knowledge representation with the Bayesian network based knowledge representation. For the development of normative expert systems which recommend an optimal decision alternative(s) based on optimization of an objective function, the use of influence diagrams [Shachter 1986, Ezawa 1986 & 1992] is a logical way to replace a heuristic based inference process of expert systems with a full fledged decision theoretic process. Evidence propagation operations (update based on new observation) on the influence diagrams are critical for the realization of the normative expert systems. Furthermore, these operations can be used for the evaluation of influence diagrams in conjunction with such as balanced sampling or simulation methods (approximal methods) [Shachter 1990]. The approximal evaluation methods are needed when influence diagrams become too large to evaluate using standard evaluation algorithms.

The value of evidence is a measure of the value of observation/experimentation. The value of evidence is based on evidence propagation operations and standard operations on the influence diagram. The need for such a measure is as follows: Now, if we want to know the impact of a variable X on Bayesian network, we often compute the entropy of X. But note that entropy is a functional of the distribution of X. It does not depend on the actual values (outcomes) of X taken by the variable X, but only on the probabilities. In the normative expert systems, we need something better than that. In decision analysis, we use such measures as the outcome sensitivity, the value of perfect information and the value of control (discussed in section 3) which reflect the actual values and probabilities of X to the objective (goal) function. Further more we want to have something finer than that of the variable level, i.e., the individual outcome level. For example, for variable X, you might be able to perform some experiments or tests to verify only selected actual values (outcomes). In this case, we want to know which particular outcomes have most significant impact to the final value (goal). That is the value of evidence. Since the value of evidence is on the outcome level, it also allows us to unify the concepts of the outcome sensitivity, the value of perfect information, and the value of control. I.e., once you know the value of evidence for each outcome, you can define these concepts from the value of evidence.

In this introduction, we briefly describe influence diagrams. We discuss evidence propagation in the presence of deterministic nodes, decision nodes, and a value node in addition to chance nodes in the regular influence diagrams in section 2. In section 3, we define and discuss the value of evidence in conjunction with the value of perfect information and the value of control. In section 4, we discuss the implementation and computational efficiency issues related to the value of evidence computation.

An influence diagram is a graphical representation of a decision problem under uncertainty, explicitly revealing probabilistic dependence and the flow of information. It is an intuitive framework in which to formulate problems as perceived by decision makers and to incorporate the knowledge of experts. It is also a mathematically precise description of the problem that can be directly evaluated.

An influence diagram is an acyclic directed graph with four types of nodes which represent different type of variables and two types of arcs which represent relationships between nodes.

A circle symbolizes a chance node which represents uncertain "event" and contains mutually exclusive potential outcomes and associated probabilities. A double circle symbolizes a deterministic node which represents functional relationships between nodes (variables) and contains a deterministic function that describes the relationships. A square symbolizes a decision node which represents a decision variable for the decision maker and contains decision alternatives. A diamond symbolizes a value node (V) which represents the goal of the decision problem and contains the value function which measures the "value."

An arc into a chance node indicates there is a probabilistic dependency between the node and its predecessor(s), and it is called a conditional arc. An arc into a decision node is an informational arc, i.e., before you make a decision, you have information related to its predecessor(s). It represents time precedence between the decision node and its predecessor(s).

A successor of node i is a node on a directed path emanating from node i. A successor node that is adjacent to node i is called a direct successor of the node i and denoted as S(i).

A predecessor of node i is a node on a directed path terminating at node i. A predecessor that is adjacent to node i is called a direct predecessor of node i and is denoted as C(i) (as conditioning nodes).

Each node i has an associated variable $X_i$, outcome space $\Omega_i$, and $x_i$ which represents a particular outcome of $\Omega_i$. A subset of $\Omega_i$ is denoted by $\mathbf{x}_i$. $\mathbf{X}$ denotes all variables, and $\mathbf{D}$ denotes all decision variables which is a subset of $\mathbf{X}$ in the influence diagram. $P\{X_i\}$ represents the probability distribution of the conditionally independent variable $X_i$. $P\{X_i|X_j\}$ represents the probability distribution of conditionally dependent variable $X_i$ given $X_j$.

There are three standard (regular) operations in the influence diagram evaluation algorithms. One is "chance node removal" which is to take the expectation of the joint probability given the chance node. The second operation is "decision node removal" which is to take the maximum (minimum) of expected value (objective function in the value node) given the alternatives of the decision node. The last operation is "arc reversal" that is to change the direction of the arc which represents an application of Bayes rule.

## 2. EVIDENCE PROPAGATION ON INFLUENCE DIAGRAMS

In this section, we discuss three types of evidence propagation, 1) chance to chance evidence propagation and evidence reversal, 2) evidence propagation with a deterministic node, and 3) evidence propagation with a decision node.

The instantiation of evidence on a chance node and propagation of evidence among chance nodes involve the following operations depending on the network structure [Shachter 1990]:

• Evidence absorption: instantiation of evidence $x_j$ on node $X_j$ which is just the table lookup of the observed outcome, i.e., $P\{X_J=x_j \mid X_{C(X_J)}\}$

• Evidence propagation: propagation of evidence $x_j$ to its successor node i, which is the identification of still valid potential outcomes,
i.e., $P\{X_i \mid X_{C(X_I)} \& X_J=x_j\} * P\{X_J=x_j \mid X_{C(X_J)}\}$

• Evidence reversal: evidence absorption of $x_j$ on $X_J$ and arc reversal between $X_J$ and its predecessor $X_K$ and the propagation of evidence $x_j$ to $X_K$.

• Evidence Propagation to a Deterministic Node:
The propagation of evidence $x_j$ to its successor node I, which is a function (F) of $X_J$ and others is to set $X_J = x_j$ in the function F., i.e., $P\{X_I = F(X_{C(I)\backslash J} \& X_J=x_j)\}$ where " \ " indicates exclusion of J.

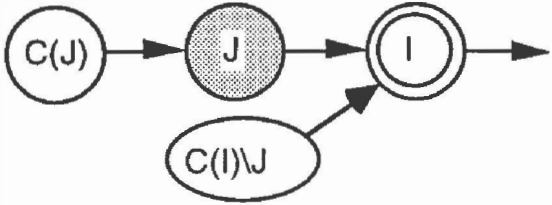

Figure 1: Evidence Propagation to a Deterministic Node

Note: We can first convert a deterministic node to a chance node, and then perform standard evidence propagation operations. An advantage of keeping it as a deterministic node is that we can save computational space better in this form in the influence diagram.

• Evidence Reversal to a Deterministic Node:
Evidence reversal to a deterministic node requires the deterministic node to be converted to a chance node first, and then perform regular (chance to chance) evidence reversal.

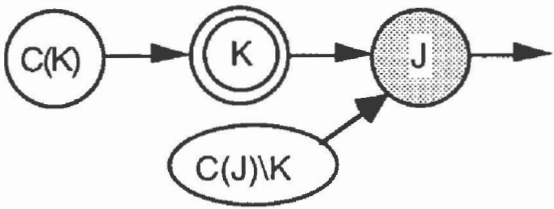

Figure 2: Evidence Reversal to a Deterministic Node

• Evidence Propagation to a Decision Node:
It involves standard evidence propagation and an elimination of an arc from J to I (decision node).

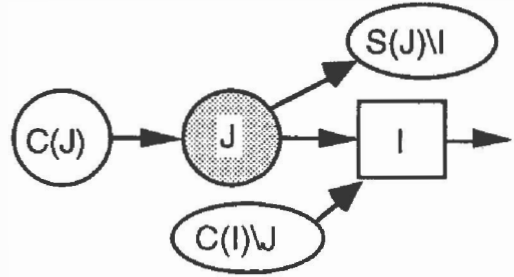

Figure 3: Evidence Propagation to a Decision Node

**Proposition 1**:
After the evidence propagation of a chance node J to its successors $S(J) \backslash I$, if the chance node has a decision successor I, we can simply delete the arc from the chance node to the decision node.

Proof:
After the evidence propagation of chance node J to all other successor nodes $S(J) \backslash I$, the decision node I becomes the only successor of J. Thus J belongs to $X_{C(I)\backslash C(V)}$ where V is a value node. In the removal of decision node I, $X_{C(I)\backslash C(V)}$ are irrelevant and do not play a role [Shachter 1986, Ezawa 1986]. Hence the arc from J to I can be eliminated. []

• Evidence propagation with a Decision Node predecessor: In this particular case of evidence propagation, there is an issue of what is the "evidence," i.e., whether we observe an outcome of a node or conditional outcomes of a node. We will discuss this issue in section 3 - the value of evidence. Here we assume we observe conditional outcomes of the node, and observation of an unconditional outcome of the node is a special case of observation of conditional outcomes.

**Proposition 2**:
A chance node with decision node predecessors can propagate evidence with observed outcomes identified as conditional i.e., J|K. With the full evidence, it contains outcomes of J given all decision alternatives of K. After the propagation of evidence of {J|K}, decision node K inherits J's successors as successors.

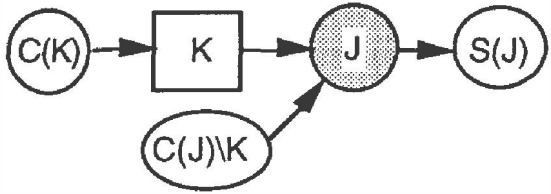

Figure 4: Evidence Propagation with a Decision Node Predecessor

Proof:
The joint probabilities, $P\{S(J) \cup J \cup C(J)\} = P\{S'(J) \cup J \cup K\}$ where $S'(J)$ represents successors of J after the arc reversals to all chance nodes of $C(J)$. If we apply evidence propagation of $J = x_j^*|x_k$, we get $S'(J = x_j^*|x_k) \cup K$. []

It is basically maintaining the dependencies between J's successors and K after the evidence propagation of J.

With these operations of evidence propagation on influence diagrams, we can now propagate evidence on the influence diagram anywhere. In the next section, we discuss the value of evidence.

## 3. VALUE OF EVIDENCE

In the process of updating influence diagrams through evidence (observation), it is often useful to know what evidence we would like to observe, or what experiment we should do to receive maximum benefit from the observation (e.g., in the case of inquiry, what specific question we should ask). We call this measure of the value of an experiment as the **value of evidence (VOE)** defined below.

$$VOE(X_J = x_j) = EV(X \setminus X_J, X_J = x_j) - EV(X)$$

i.e., the value of evidence of outcome $x_j$ of variable $X_J$ is the expected value (EV) of the influence diagram with the evidence $x_j$ minus the expected value of the influence diagram. Because the influence diagrams have value node (value function), we can compute this value of evidence based on the value function of the value node.

In the normative expert systems which are based on influence diagram paradigm, the evidence propagation operation is the most often used functionality. It would be convenient if we could compute the value of perfect information and the value of control from the measure based on evidence propagation. The value of evidence is the measure which allows us to compute the value of perfect information and the value of control. Outcome sensitivity can also be computed from the value of evidence.

The outcome sensitivity determines changes in expected value of a given influence diagram and takes into account all uncertainty of other chance nodes given a specific outcome of a node. The **outcome sensitivity (OS)** is defined by

$$OS(X_J) = \text{Max } EV(X \setminus X_J, X_J=x_i) - \text{Min } EV(X \setminus X_J, X_J=x_j) \text{ for all } \Omega_J.$$

We can also define the outcome sensitivity by

**Proposition 3:**
$$OS(X_J) = \text{Max VOE}(X \setminus X_J, X_J=x_i) - \text{Min VOE}(X \setminus X_J, X_J=x_j) \text{ for all } \Omega_J.$$

Proof:
$OS(X_J)$ = Max $EV(X \setminus X_J, X_J=x_i)$ - Min $EV(X \setminus X_J, X_J=x_j)$ for all $\Omega_J$.
= Max $(EV(X \setminus X_J, X_J=x_i) - EV(X))$ - Min$(EV(X \setminus X_J, X_J=x_j) EV(X))$ for all $\Omega_J$.
= Max $VOE(X \setminus X_J, X_J=x_i)$ - Min$VOE(X \setminus X_J, X_J=x_j)$ for all $\Omega_J$. []

I.e., the outcome sensitivity is the difference between the maximum value of the value of evidence and the minimum value of the value of evidence on the variable $X_J$.
This particular sensitivity is different from so called "deterministic sensitivity" which set all other chance node to the median.

The value of perfect information is the difference between the expected values of knowing the outcomes of a node before making a decision and not knowing the outcome of the node of a given influence diagram in question. The **value of perfect information (VOPI)** is defined as

$$VOPI(X_J) = EV(X \setminus \{D, X_J\}, D|X_J, X_J) - EV(X).$$

In the influence diagram operation, to obtain the expected value with perfect information, we add an arc from the chance node in question to the related decision node. The difference in the expected values of this modified influence diagram and of the original influence diagram is the value of perfect information. We call this "standard method" of the computation of the value of perfect information. When a decision node is a predecessor, we perform a special operation which we will discuss later.

We can also define the value of perfect information as the expected value of the value of evidence of the evidence node J, i.e.,

**Proposition 4:**
VOPI $(X_J) = \Sigma$ VOE$(X_J = x_j) * P\{x_j\}$ for $\Omega_J$ of evidence node j.

Proof:
VOPI$(X_J)$=EV$(X\backslash\{D,X_J\}, D|X_J, X_J)$-EV$(X)$
=$(\Sigma$EV$(X\backslash\{D,X_J\},D|X_J,X_J=x_j)*P\{x_j\})$- EV$(X)$
=$(\Sigma$EV$(X\backslash X_J,X_J=x_j)*P\{x_j\})$-EV$(X)$
=$\Sigma$(EV$(X\backslash X_J, X_J=x_j)$-EV$(X))*P\{x_j\}$
= $\Sigma$ VOE$(X_J = x_j) * P\{x_j\}$
for $\Omega_J$ of evidence node j. []

In other words, once the evidence $x_j$ is propagated, when we make the next decision (reduce decision node), this information is already incorporated(i.e., given the evidence of $x_j$.) Hence by weighing the value of evidence for each $x_j$ with $P\{x_j\}$, we can compute the value of perfect information. The unconditional probability $P\{X_J\}$ can always be obtained by applying arc reversals between its predecessors as long as they are not decision nodes. In the case of decision predecessors, the treatment of the probability is discussed in a example later.

Note that VOPI computed from VOE is the VOPI for overall decisions. Only when a decision predecessor is involved, we can compute VOPI for the decision. Note also a value of evidence could be negative, but the value of perfect information is always greater than or equal to 0.

The value of control is the difference between the expected values of controlling the outcomes of the node in question and not controlling it. The **value of control (VOC)** is defined by

VOC$(X_J)$ = Max EV$(X\backslash X_J, X_J=x_J)$ - EV$(X)$ for all $\Omega_J$.

In the influence diagram operation, to obtain the expected value with control, we change the chance node in question to a decision node.

The value of control can be directly computed from the value of evidence.

**Proposition 5:**
VOC $(X_J)$ = Max VOE $(X_J = x_j)$ for $\Omega_J$ of evidence node J, if we are maximizing the value function.

Proof:
VOC$(X_J)$=Max EV$(X\backslash X_J,X_J=x_j)$ - EV$(X)$
= Max (EV$(X\backslash X_J, X_J=x_j)$ - EV$(X)$ )
= Max VOE $(X_J = x_j)$ for $\Omega_J$ of evidence node J []

In other words, the evidence absorption is controlling of the event, i.e., assuming certainty of the outcome. Choosing the best outcome is the value of control. Hence by choosing the value of evidence which optimizes the value function, we get the value of control.

Now let's discuss the issue of evidence/observation of conditional outcomes and unconditional outcomes in the case of evidence propagation with a decision node predecessor. We discuss this in conjunction with the value of evidence, the value of perfect information, and the value of control using an example, Mars vs Venus [Matheson, 1990].

**Mars vs Venus:** Consider an hypothetical case of sending a landing craft to Mars or Venus. As a decision maker, we have a choice of sending the craft to Mars or Venus. The probability of success of the mission is 60% regardless of the destination. The values we receive from the mission is as follows:

Table 1: Mars vs Venus

| Destination | Mission (Probability) | Value |
|---|---|---|
| Mars | Success (60%) | 50 |
| Mars | Failure (40%) | 10 |
| Venus | Success (60%) | 100 |
| Venus | Failure (40%) | -10 |

Figure 5 shows an influence diagram representation of Mars vs Venus problem. Note that an arc from the decision node (Destination) to the chance node (Mission) is not necessary, since the probability of success or failure is independent of the destination. But for the sake of the consistency with the original example, we avoid modification to the diagram nor the changes in probabilities.

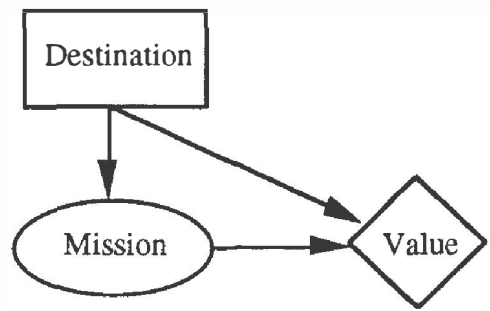

Figure 5: Mars vs Venus

The expected value of this decision problem is 56, and optimal destination is Venus. The value of evidence for the mission is as follows:

Table 2: Value of Evidence

| Evidence | Destination | Value | VOE |
|---|---|---|---|
| Failure | Mars | 10 | - 46 |
| Success | Venus | 100 | + 44 |

The expected value with perfect information is 64, thus the value of perfect information is 8. It is computed from the value of evidence using information from table 1 and 2 (i.e., -46 * 0.40 + 44 * 0.60 = 8.) The value of control is 44 (i.e., the maximum value of the value of evidence.) This is a "naive" computation of value of perfect information (i.e., based on observation of unconditional outcomes.) Next, we discuss a more sophisticated interpretations of the value of perfect information (i.e., based on an observation of conditional outcomes.)

### 3.2    With Full Evidence

In the case of value of perfect information, if we don't assume conditional independence between probability of success in Mars landing and Venus landing, we need to reassess these conditional probabilities for the computation of the value of perfect information for both using and not using value of evidence. For the computation of the value of perfect information, since we cannot directly reverse the arc between the "destination" and the "mission" due to the fact that the "destination" is a decision node in Figure 5, we need to modify the influence diagram to the one in Figure 6, which allows us to add an arc to Destination from Mission given Destination. Note that as shown in section 2, the value of evidence doesn't require this structural modification of the influence diagram, and the necessary change in outcomes and probability distributions can be accommodated internally in "mission" node.

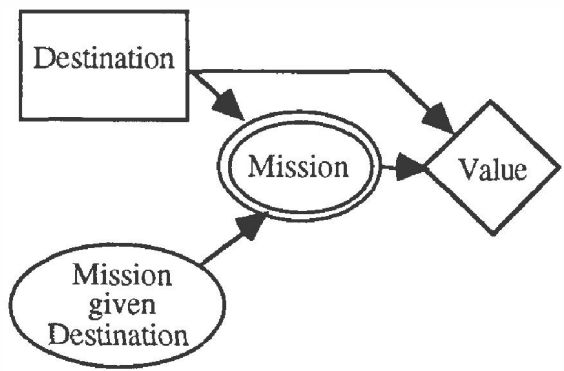

Figure 6: Modified Mars vs Venus Influence Diagram

Table 3: The joint probability distribution for the conditional outcomes of Mission given Destination

|  | Venus: Failure | Venus: Success | Sum |
|---|---|---|---|
| Mars: Failure | 0.354 | 0.046 | 0.40 |
| Mars: Success | 0.046 | 0.554 | 0.60 |
| Sum | 0.40 | 0.60 |  |

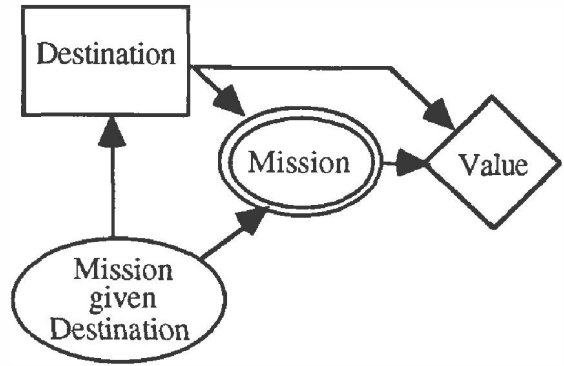

Figure 7: Mars Vs Venus with Value of Perfect Information on Mission given Destination

The expected value is 56, and optimal destination is Venus. The expected value of this mission with perfect information is 65.84, and thus the value of perfect information is 9.84.    For the value of evidence for the conditional observation, is as follows:

Table 4: Value of Evidence

| Evidence (Observation) | Desti-nation | Value | VOE |
|---|---|---|---|
| Mars: Failure Venus: Failure | Mars | 10 | - 46 |
| Mars: Success Venus: Failure | Mars | 50 | - 6 |
| Mars: Failure Venus: Success | Venus | 100 | + 44 |
| Mars: Success Venus: Success | Venus | 100 | + 44 |

As in the previous example, the value of perfect information can be directly computed from table 3 and 4. The direction of the value of evidence, i.e., + or -, is also informative, it shows the direction of expected value's change as we observe more evidence. The value of control remains 44.

### 3.3    With Partial Evidence

In the case with partial evidence for both the value of perfect information and the value of evidence require further assessment of conditional probability among outcomes, since we observe only part of evidence (conditional outcome). Figure 8 shows an influence diagram which separates Mars Landing and Venus Landing.

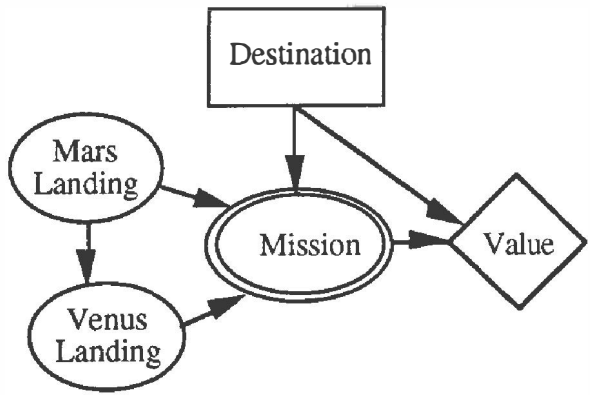

Figure 8: Modified Mars Vs Venus Influence Diagram with Dependency between Mars and Venus Landings

The table 5 shows the probability distribution for Venus Landing which is derived from Table 4.

Table 5: The probability distribution for the conditional outcomes of
Venus Landing given Mars Landing

|  | Venus Landing: Failure | Venus Landing: Success | Unconditional Prob. |
|---|---|---|---|
| Mars Landing: Failure | 0.885 | 0.115 | 0.40 |
| Mars Landing: Success | 0.077 | 0.923 | 0.60 |
| Unconditional Prob. | 0.4 | 0.6 |  |

The value of Evidence on Mars Landing is as follows:

Table 6: The Value of Evidence for Mars Landing

| Evidence | Destination | Value | Value of Evidence |
|---|---|---|---|
| Mars Landing: Failure | Mars | 10 | -46 |
| Mars Landing: Success | Venus | 91.57 | 35.57 |

The value of perfect information is 2.94 (i.e., -46 * 0.40 + 35.57 * 0.60), and value of control is 35.57.

In this section, we defined and discussed the concept of value of evidence and relationships between the value of evidence and the outcome sensitivity, the value of perfect information, and the value of control.

The advantage of the use of value of evidence is that we can avoid the unnecessary modification of influence diagrams like the one we discussed here. Also it allows us to compute the outcome sensitivity, the value of perfect information and the value of control directly from the value of evidence. As we discussed in the next section, it is also computationally more efficient in certain applications.

## 4    IMPLEMENTATION AND COMPUTATIONAL ISSUES

In this section, we discuss two different methods to compute the value of evidence, and discuss a relative computational efficiency.

Two methods for the computation of the value of evidence are as follows:

**Method 1 (based on Evidence Propagation):**

For an outcome, $x_i$, of an evidence node I,
• Perform evidence propagation of $x_i$,
• Perform standard reduction of influence diagram,
• Compute the value of evidence of $x_i$,
  Repeat this procedure for all $x_i$ of I.

**Method 2 (based on Lock Selected Node):**

• Lock an evidence node I,
• Perform standard reduction of influence diagram,
• Compute the value of evidence of all $x_i$ of I.

In method 2, "lock a node I" means that we put the node I to be non-removable node. If we perform standard reduction on the influence diagram, we get the expected values conditional on the outcomes ($x_i$s) of the locked node I. The value of evidence $x_i$ is this conditional expected value of $x_i$ minus the expected value without evidence (as defined in the previous section.)

The method 1 computes a value of evidence one outcome at a time, whereas the method 2 computes all the values of evidence at once.

Now, we compare the computation of the value of perfect information, the value of evidence using method 1 and 2. We ignore the computational time and space required for the conversion of the values of evidence to the value of perfect information. We show that the computation of the value of perfect information using the standard method and method 2 are equivalent in terms of maximum computational outcome space, and method 1 is more efficient than or at least equal to those two.

In the reduction (evaluation) of influence diagram, one of the most critical space limitation is the maximum computational outcome space requirement. The computational outcome space of a reduction of node J is $\Omega_{S(J) \cup J \cup C(J) \cup C(S(J))}$. So the bottleneck of the evaluation of influence diagram is Max. $\Omega_{S(J) \cup J \cup C(J) \cup C(S(J))}$ for all J in the sequence of the reduction of influence diagram. I.e., if the maximum computational outcome space exceeds the available space in the computer (hardware), it simply stops and won't be able to evaluate the influence diagram in question. We need to be able to evaluate the influence diagram before we worry about how fast we can evaluate it.

Figure 9 shows an example influence diagram with an evidence node I and a decision node K.

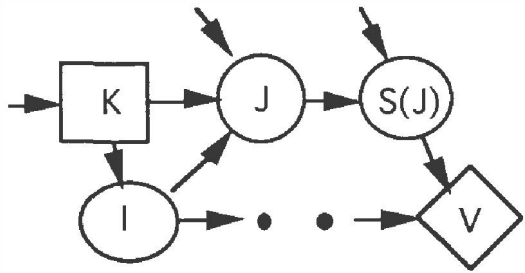

Figure 9: Original Influence Diagram

Figure 10 shows a modified influence diagram for the computation of the value of perfect information.

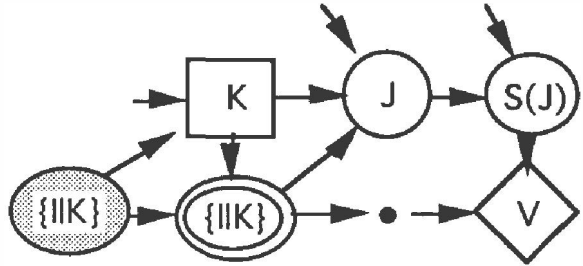

Figure 10: Influence Diagram for Value of Perfect Information

Figure 11 shows a modified influence diagram for the computation of the value of evidence.

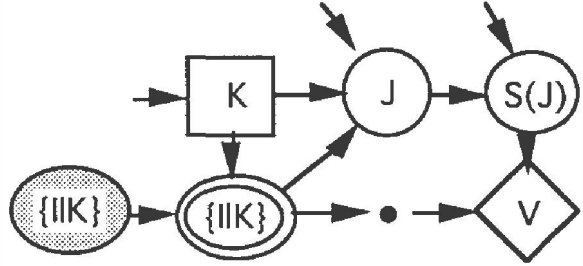

Figure 11: Influence Diagram for Value of Evidence

**Proposition 6:**
The maximum computational outcome space of the value of perfect information using the standard method and the method 2 are equivalent.

Proof:
If a reducing node $J^*$ is not a successor of node $\{I|K\}$ in the sequence of reduction of the influence diagram and requires maximum computational outcome space, then both computation require the same maximum outcome space. In the case of the reducing node $J^*$ to be a successor of node $\{I|K\}$, since until the reduction of decision node K, both follow the same order of reduction of nodes, both computation require the same maximum outcome space. []

**Proposition 7:**
The maximum computational outcome space of the value of perfect information using standard method is larger than or equal to that of using method 1.

Proof:
If a reducing node $J^*$ is not affected by the evidence propagation of node $\{I|K\}$ and requires maximum computational outcome space in the sequence of reduction of the influence diagram, then both computation require the same maximum outcome space. But if the node $J^*$ is impacted by the evidence propagation of node $\{I|K\}$, since arc reversal operation and evidence propagation operation requires one variable less (i.e., $\{I|K =x_i|x_k\}$), the maximum computational outcome space requirement for the reduction of node $J^*$ is less than that of the standard method. []

Hence in terms of the maximum computational requirement, the method 1 is the most efficient way to compute the value of evidence, and the value of perfect information. This also holds for the computation of the outcome sensitivity and the value of control.

## 5  SUMMARY

Evidence propagation operations with deterministic, decision, and value nodes on influence diagrams and the value of evidence are discussed. These operations are crucial to use influence diagrams in the normative expert systems for the general update based on new evidence/observation. The value of evidence is useful to measure the value of experimentation. We can compute the value of perfect information and value of control from the values of evidence. We also discussed that the efficient methods for the computation of value of evidence.